\documentclass{article}
\pdfoutput=1
 
\usepackage[letterpaper,top=2cm,bottom=2cm,left=3cm,right=3cm,marginparwidth=1.75cm]{geometry}

\usepackage{amsmath}

\usepackage{graphicx}
\usepackage[colorlinks=true, allcolors=blue]{hyperref}

\usepackage{tabu}                     
\usepackage{
multirow}                 
\usepackage{multicol}                 
\usepackage{multirow}                
\usepackage{float}                    
\usepackage{makecell}                 
\usepackage{booktabs}                 
\usepackage[normalem]{ulem}

\useunder{\uline}{\ul}{}
\usepackage{graphicx}
\usepackage{placeins} 

\usepackage{abstract} 
 
\usepackage{amsmath} 
\usepackage{bbding}

\newcommand\keywords[1]{\textbf{keywords}: #1} 

\title{MOT\_FCG++: Enhanced Representation of Spatio-temporal Motion and Appearance Features}
\author
{Name:Yanzhao Fang\\email:fangych7@mail2.sysu.edu.cn} 
 
\begin{document} 
\maketitle

\begin{abstract}
The goal of multi-object tracking (MOT) is to detect and track all objects in a scene across frames, while maintaining a unique identity for each object. Most existing methods rely on the spatial-temporal motion features and appearance embedding features of the detected objects in consecutive frames. Effectively and robustly representing the spatial and appearance features of long trajectories has become a critical factor affecting the performance of MOT. We propose a novel approach for appearance and spatial-temporal motion feature representation, improving upon the hierarchical clustering association method MOT\_FCG. For spatial-temporal motion features, we first propose Diagonal Modulated GIoU, which more accurately represents the relationship between the position and shape of the objects. Second,  Mean Constant Velocity Modeling is proposed to reduce the effect of observation noise on target motion state estimation. For appearance features, we utilize a dynamic appearance representation that incorporates confidence information, enabling the trajectory appearance features to be more robust and global. Based on the baseline model MOT\_FCG, we have realized further improvements in the performance of all. we achieved 63.1 HOTA, 76.9 MOTA and 78.2 IDF1 on the MOT17 test set, and also achieved competitive performance on the MOT20 and DanceTrack sets.

\end{abstract} 


\begin{figure}[htbp]
\centering
\includegraphics[width=1.0\linewidth]{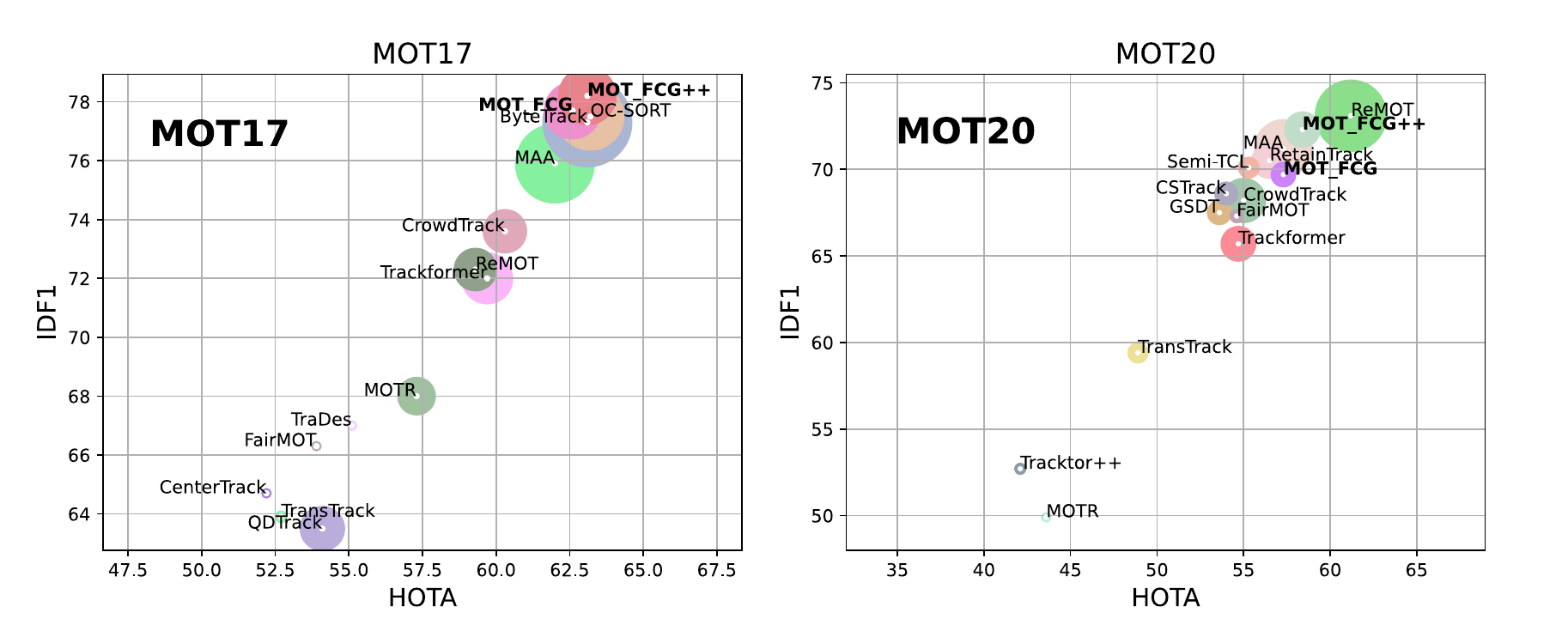}
\caption{\label{fig:MOT17} Comparison of Different Trackers on the MOT17-test Set and MOT20-test Set in Terms of IDF1, HOTA, and MOTA. The horizontal axis represents HOTA, while the vertical axis represents IDF1. The radius of the circles corresponds to MOTA. Our method, MOT\_FCG++, achieves 76.9 MOTA, 78.2 IDF1 and 63.1 HOTA on the MOT17-test set and 68.1 MOTA, 72.3 IDF1 and 58.4 HOTA on the MOT17-test set, demonstrating strong competitiveness. Please refer to the table \ref{tab:w1} for further details.}
\end{figure}

\section{Introduction} 
Multi-object tracking (MOT) aims to identify dynamic objects in each frame of a given video sequence and assign the same identity identifier to the same object across consecutive frames, thereby forming individual motion trajectories for different objects. The goal of MOT is to simultaneously track and recognize multiple targets in a scene, demonstrating enormous potential in fields such as video surveillance, intelligent transportation, autonomous driving, and sports broadcasting.

Detection-based tracking methods (DBT) are mainstream methods for MOT, typically divided into two stages: detection and tracking. In the first stage, a convolutional neural network-based detector is used to label the objects of interest in each frame of the video sequence with bounding boxes. In the second stage, the tracker utilizes the output from the detector to extract the appearance and spatial motion features of the objects again, and calculates the similarity between these detection features and the existing trajectory features to perform trajectory association. 

MOT\_FCG\cite{girbau2022multiple} proposed recently by Andreu Girbau is one of the DBT paradigm tracking methods. MOT\_FCG utilizes the idea that temporally similar target instances are similar in appearance to form an initial trajectory, and constructs the final target trajectory by incorporating spatio-temporal motion coherence constraints in a hierarchical clustering fashion. However, we found that using IoU for spatio-temporal motion feature representation in MOT\_FCG is inaccurate. Additionally, using the median element for appearance feature representation in long trajectories can introduce low-quality appearance features, which inevitably leads to biases during trajectory association.

Therefore, how to effectively represent the appearance and spatio-temporal motion features of trajectories is a key factor affecting the performance of DBT methods. In this paper, we improve upon the hierarchical clustering tracking method MOT\_FCG by proposing a more global appearance embedding representation method and a more precise spatio-temporal motion information representation method. Ablation experiments validate the effectiveness of our proposed modules, and as shown in Figure \ref{fig:MOT17} we achieve excellent performance on the MOT17 and MOT20 datasets.
 
The main contributions of this paper are threefold:
\begin{itemize}
\item[$\bullet$] We point out that the use of median elements as trajectory appearance embedding features in MOT\_FCG has limitations. To address this, we adopt a dynamic appearance embedding representation method. This method can adaptively adjust the weighting based on confidence. It integrates the global appearance embedding features of the trajectory, leading to a more comprehensive and holistic representation.
\item[$\bullet$] To more accurately represent spatio-temporal motion information, we propose a new spatial metric method—Diagonal Modulated GIoU and an improved motion state estimation model-Average Constant Velocity Modeling. This method can more precisely characterize the positional relationships between objects and to some extent reflect the shape information of the object bounding boxes. 
\item[$\bullet$]  We achieved competitive results on the MOT17 and MOT20 datasets, and the results show that, compared to the baseline model, all metrics have improved. We also validated the effectiveness of the proposed modules on the DanceTrack dataset.
\end{itemize}

\section{Related Work}
The main aspect of MOT is the data association between current frame detections and tracking trajectories. Based on the criteria for this association, MOT can be primarily divided into motion-based MOT and appearance-based MOT. The former is efficient in terms of run time but is easily affected by occluded objects; the latter achieves better tracking performance but always has lower operational efficiency. Recent research on MOT mainly includes the following: 

\textbf{Motion-Based Multi-Object Tracking.} Detection-based trackers often perform frame-by-frame association by linking detected objects with trajectory motion and positional cues. Within adjacent frames, the displacement of objects is typically small and can be approximated as linear motion. Therefore, the Kalman filter \cite{brown1997introduction} can be used to predict the object positions using a constant velocity model. For the representation of motion features, the primary considerations are motion prediction compensation and the relationships between object positions. SORT \cite{bewley2016simple} employs a Kalman filter to predict the next state of the observations provided by the detector and performs MOT based on Intersection over Union (IoU). BoT-SORT \cite{aharon2022bot} considers the influence of camera motion on linear predictions and employs motion compensation to correct the predictions of the Kalman filter. OC-SORT 
 \cite{cao2023observation}reactivates lost observations and backtracks over the periods of missing observations to update the Kalman filter parameters. Strong-SORT \cite{
du2023strongsort} applies offline correction to achieve Gaussian smoothing of the predicted trajectories. Recent advancements have introduced Transformer \cite{2017Attention} that follow the attention tracking paradigm. These methods utilized complex motion models \cite{2020TransTrack,2021TrackFormer,2021TransCenter,2022MOTR,2021TransMOT} and have achieved outstanding performance across several datasets.   
 
\textbf{Appearance-Based Multi-Object Tracking.} To achieve better performance in long-term association scenarios, many methods utilize additional re-identification networks to encode the appearance of objects, which serve as cues for matching appearance features. DeepSORT \cite{2017Simple} is one of the earliest methods to use appearance features for association. Tracktor++ \cite{2019Tracking} utilizes Faster R-CNN \cite{2017Faster} for regression prediction of trajectory positions and associates inactive trajectories with detections through a simple ReID network. CTracker \cite{2020Chained} extends single-frame regression to pairwise regression across adjacent frames, incorporating a ReID module as identity attention, thereby forming an end-to-end chain structure for detection and tracking. For trajectory feature representation, JDE \cite{2020Towards} employs an Exponential Moving Average (EMA) to update trajectory appearance features, while Deep-OC-SORT \cite{2022Observation} integrates confidence information to correct the EMA.

\textbf{Multi-Object Tracking Paradigms.} Deep learning-based MOT algorithms mainly include detection-based tracking (DBT) and joint detection and tracking (JDT). DBT primarily consists of SORT-like methods \cite{bewley2016simple,aharon2022bot,cao2023observation,2017Simple} and graph convolutional network methods \cite{2020Graph, Bras2019Learning}. JDT mainly includes methods based on Transformers \cite{2017Attention,2020TransTrack,2021TrackFormer,2021TransCenter,2022MOTR}, Siamese networks \cite{2019How,2019Tracking, 2021SiamMOT}, and deep feature joint reuse \cite{2021FairMOT,2020Towards, 2020RetinaTrack}. 

The clustering tracking association is a relatively novel DBT method, such as MOT\_FCG \cite{girbau2022multiple}, which clusters detections based on the similarity of appearance features from adjacent frames to form different tracklets. It incorporates spatio-temporal information modulation into the appearance features. However, we observe certain limitations in the representation of appearance features and spatial motion during the association process. Inspired by Hybrid-SORT \cite{yang2024hybrid}, we introduce subtle shape and confidence information into the model. Although these are relatively weak cues, they will play an important role in clustering association.

This paper uses MOT\_FCG as the baseline, introducing subtle confidence and object shape information to achieve a better representation of object motion and appearance. This enhancement provides clustering tracking methods with greater robustness and generalization capability, and we hope it inspires subsequent tracking algorithms.

\section{Method}
\subsection{Preliminary: MOT\_FCG}
\begin{figure}[htbp]
\centering
\includegraphics[width=0.5\linewidth]{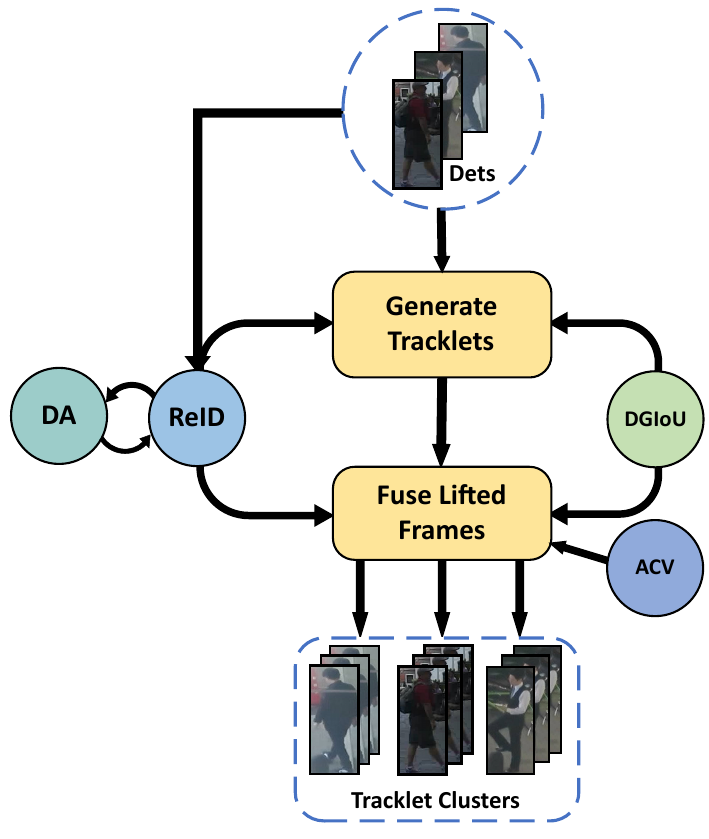}
\caption{\label{fig:MOT_FCG++}Illustration of MOT\_FCG++}
\end{figure}

Our work is based on the recently proposed hierarchical clustering tracking algorithm MOT\_FCG \cite{girbau2022multiple}. Because the same object instance exhibits similar appearance features in close temporal neighborhoods, MOT\_FCG utilizes the similarity of objects in adjacent frames to generate clustered tracklets. The algorithm comprises two main stages. In the first stage, initial, short and reliable tracklets, referred to as lifted frames, are formed based on appearance features within a specified time window. In the second stage, the authors use the appearance features of the median elements of these tracklets as representations. MOT\_FCG continuously refine the hierarchy of lifted frames through a hierarchical clustering algorithm, which naturally merges the tracklets to create the final target trajectory. The authors implemented UPGMA (Unweighted Pair Group Method with Arithmetic Mean) \cite{hibbert2017unweighted} to iteratively merge paired clusters \cite{2020Author}, creating a hierarchical structure. They also add spatio-temporal motion features to aggregate the tracklets with spatio-temporal coherence and appearance similarity as much as possible. As shown in Figure \ref{fig:MOT_FCG++}, we propose MOT\_FCG++ based on MOT\_FCG, introducing three modules: Diagonal Modulated GIoU, Dynamic Appearance, and Average Constant Velocity Modeling to represent weak cue information, thereby enhancing the robustness and generalization of the clustering association model.

\subsection{Clustering Spatial Motion Features Representation}
\begin{figure}[htbp]
\centering 
\includegraphics[width=0.6\linewidth]{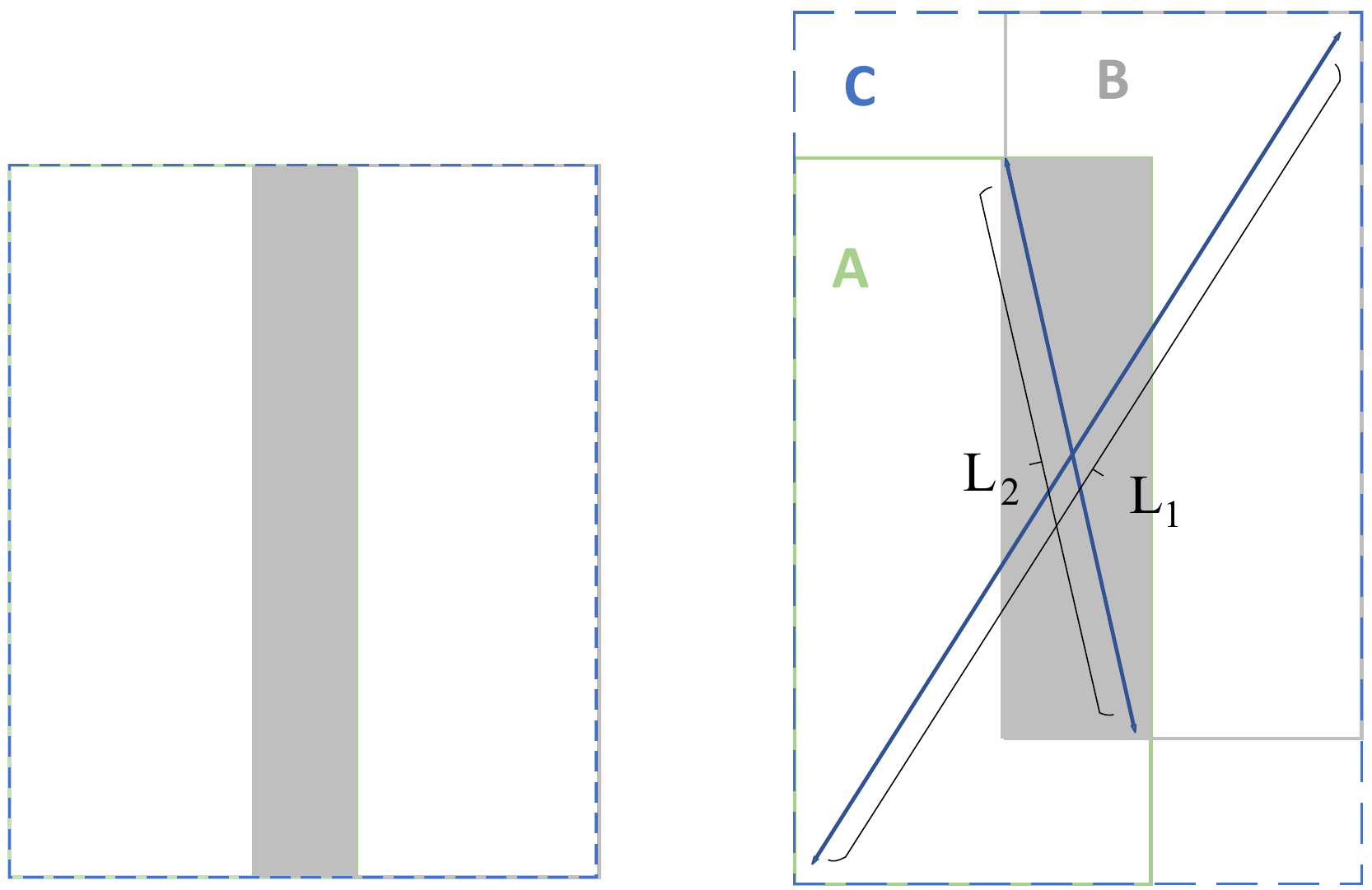}
\caption{\label{fig:DGIoU}Comparison between IoU and Diagonal Modulated GIoU}
\end{figure}

In the association between current frame detections and tracking trajectories, IoU is often calculated to represent the spatial motion information of objects. As shown in Figure \ref{fig:DGIoU}, IoU is defined as the area of intersection of two detection bounding boxes divided by the area of their union. However, this approach has two main issues: first, IoU does not accurately reflect the degree of overlap between the two boxes. Figure \ref{fig:DGIoU} clearly shows that, the overlap on the left is greater than that on the right, even when the IoUs are equal. Second, IoU is scale-invariant and fails to accurately represent the shape relationship between two objects. Besides,methods based on clustering association lack the assistance of Kalman prediction for spatial motion information, necessitating a more accurate spatial position metric. Therefore, to address these two issues, we propose a spatial position metric called Diagonal Modulated GIoU as a substitute:

\begin{equation}
\begin{aligned}\begin{array}{ccc}
     & \mathrm{GIoU}=\frac{|(A\cap\mathrm{B})|}{|(A\cup\mathrm{B})|}-\frac{|C-(A\cup\mathrm{B})|}{|C|},\\[10pt]
     & d_{\mathrm{DGIoU}}=1-\frac{\mathrm{L}_{2}}{\mathrm{L}_{1}}\cdot\mathrm{GIoU}, \\[10pt]
     & \lambda_{\mathrm{C}}=\min\left\{1,\frac{d_{\mathrm{DGIoU}}}2+\mathrm{off}\right\},\\
     \end{array} \end{aligned}
\end{equation} 
where $A$ and $B$ represent the areas of two objects, and $C$ denotes the area of the smallest rectangle that can encompass both $A$ and $B$. It is important to note that dividing $d_{\mathrm{DGIoU}}$ by 2 is done for normalization, ensuring that $d_{\mathrm{GIoU}}\in[0,1]$. GIoU can more accurately reflect the positional relationships between objects. Inspired by Hybrid-SORT \cite{yang2024hybrid}, we use the $\frac{\mathrm{L}_{2}}{\mathrm{L}_{1}}$ term to modulate the diagonal GIoU, which can first compensate for GIoU's resolution regarding target positional relationships. Additionally, $\mathrm{L}_{1}$ and $\mathrm{L}_{2}$ are related to the aspect ratios of the bounding boxes, to some extent, reflecting the shape information relationship between the objects.

\subsection{Clustering Appearance Information Representation}
For each tracklet of lifted frames, accurately representing their appearance features is key to clustering association. For the tracklet $T_{[n,m]}^{k}$ formed by clustering, where $n$ and $m$ are the starting and ending frame indices of the tracklet, MOT\_FCG uses the appearance features of the median element, specifically the ${\frac{n+m}{2}}^{th}$ element of the tracklet $T_{[n,m]}^{k}$, as the representation of clustering appearance features. This approach has two issues: first, it lacks global appearance information for the cluster, making the representation not robust or global. When the tracklet is longer, the appearance feature of the median element is weaker in the time continuity of data association, which will greatly affect the accuracy of hierarchical clustering association. Second, as shown in Figure \ref{tab:xiao}, MOT\_FCG does not take into account the influence of detection quality on the appearance feature representation of the trajectory. Due to environmental influences such as occlusion or background interference, it is easy to select appearance feature vectors with poor quality and interference information by using only the median element, which cannot accurately reflect the appearance feature of the trajectory.
  
\begin{figure}[htbp] 
\centering 
\includegraphics[width=0.9\linewidth]{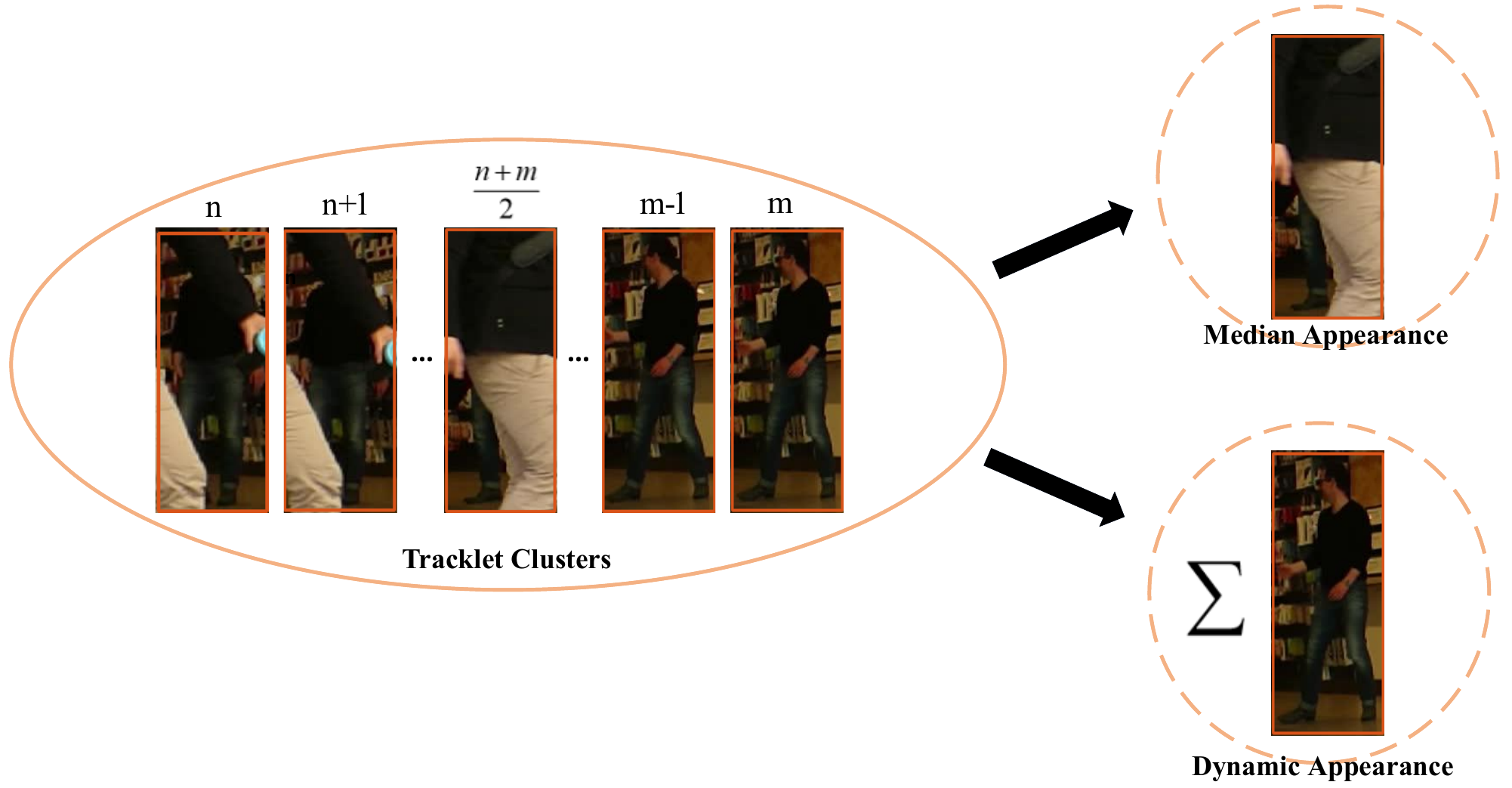}
\caption{\label{fig:DA} Limitations of Median Feature Representation}
\end{figure}

We refer to the representation method for Dynamic Appearance in Deep-OC-SORT \cite{2022Observation}, which describes the deep visual embedding for trajectories using the per-frame exponential moving average (EMA) of deep detection embeddings, along with an adaptive weight. This adaptive weight allows for a higher proportion of appearance information to be combined in high-quality situations, while a lower proportion is combined in low-quality situations. Therefore, it can adaptively recognize image degradation caused by occlusion or blurriness, while selectively rejecting corrupted embeddings. The standard EMA is:
\begin{equation}
\mathbf{e}_t=\beta\mathbf  
{e}_{t-1}+(1-\beta )\mathbf{e}^{\mathrm{new}}
\end{equation} 
where $\mathbf{e}_t$ is the appearance feature of the newly matched detection added to the lifted frame, and the adaptive weight $\beta_t$ is:
\begin{equation}
\beta_t=\beta_f+(1-\beta_f)(1-\frac{s_{\det}-\sigma}{1-\sigma}) 
\end{equation} 
where $s_{\det}$ is the detection confidence, and $\sigma$ is the confidence threshold used to filter out noisy detections. When $s_{\det} < \sigma$, it is considered a noisy detection and does not participate in the weighting of appearance embeddings; when $s_{\det} \geq \sigma$, as $s_{\det}$ increases, $\beta_t$ decreases, leading to a higher proportion of weighting for the new appearance embedding. When $s_{\det} = 1$, the maximum amount of new appearance embedding can be added. Using dynamic appearance information representation allows for linking the global appearance features of tracklets while controlling the weighting proportion with target confidence, thereby better representing the clustering appearance.
  
\subsection{Average Constant Velocity Modeling} 
MOT\_FCG integrates a simple motion estimation by calculating the difference between the position coordinates of the bounding boxes in the $t^{th}$ frame and the $(t-1)^{th}$ frame to estimate velocity, which is then applied to the bounding box position estimation in the $(t+p)^{th}$ frame. However, this approach does not account for the effects of inter-frame spacing and noise. We modify this by averaging the velocity over the last $N$ frames, using the average velocity as a constant velocity estimate. The advantages of this method are twofold: first, it reduces the impact of observational noise on velocity calculation. Secondly, considering the distance between the starting and ending frames of the trajectory aligns better with the object displacement patterns. Therefore, the average constant velocity $v$ is modeled as:
\begin{equation}
\begin{aligned}&v=\frac{
\left(x_t-x_{t-N}\right)}{N},\\&x_{t+p}=x_t+v\times p\end{aligned}
\end{equation}
where $v$ is the average velocity over the last $N$ frames, $x_t$ is the target coordinate position at frame $t$, and $x_{t+p}$ is the predicted target coordinate position at frame $t+p$. It is worth noting that if the trajectory length is less than n, the average velocity of the current trajectory is used instead. The value of n will be analyzed in the ablation study in the next chapter.

\section{Experiments} 
\subsection{Experimental Settings}
\begin{figure}[htbp]
\centering

\includegraphics[width=0.9\linewidth]{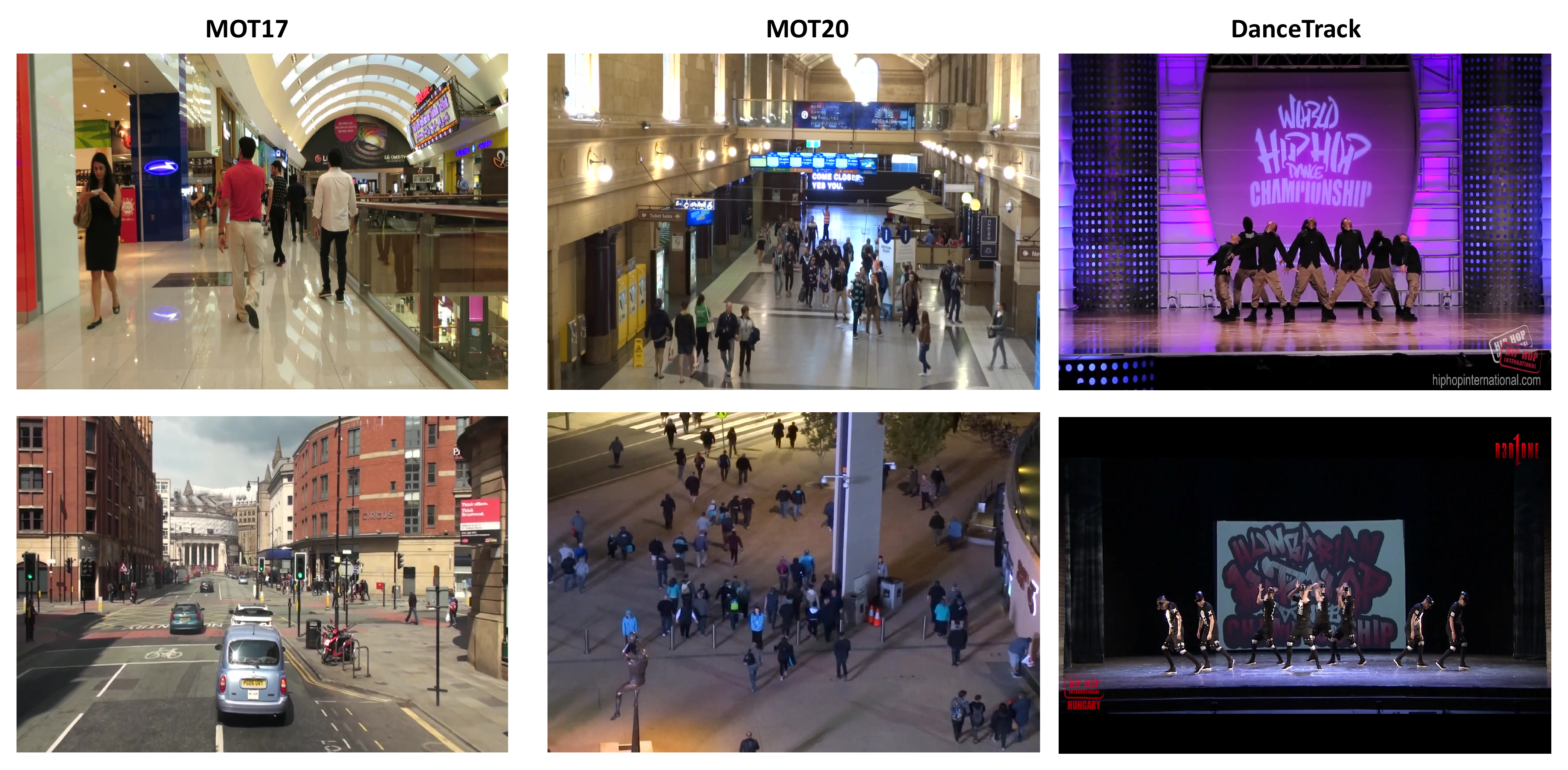}
\caption{\label{fig:datasets}Illustrations of MOT17, MOT20, and DanceTrack}
\end{figure}

\textbf{Datasets.} We conduct experiments on multiple datasets to ensure the universality and robustness of the proposed method. The datasets include the currently popular pedestrian tracking datasets MOT17 \cite{2016MOT16}, MOT20 \cite{2020MOT20}, and DanceTrack \cite{0DanceTrack}. As shown in Figure \ref{fig:datasets}, MOT17 is a widely used standard benchmark in MOT, where the motion is predominantly linear. In contrast, MOT20 features higher density and longer sequences, designed to evaluate tracking performance in scenarios with dense objects and severe occlusions. DanceTrack is one of the most challenging benchmarks in the field of MOT, characterized by diverse nonlinear motion patterns, frequent interactions and occlusions, as well as challenges posed by dancers wearing similar outfits.
 
\textbf{Metrics.} We use the CLEAR MOT \cite{2008Evaluating} metrics, including Multiple Object Tracking Accuracy (MOTA), Recall, False Positives (FP), False Negatives (FN), and ID Switches (IDSW) and so on. MOTA is calculated based on FP, FN, and IDSW, with a greater emphasis on detection performance. We also utilize the IDF1 metric, which evaluates identity association performance and reflects the accuracy of the tracker’s associations \cite{2016Performance}. Additionally, we consider the HOTA \cite{Luiten2021HOTA} metric, which clearly integrates precise detection, association, and localization effects into a single measurement to assess the overall performance of the algorithm.

\textbf{Implementation details.} We ran a test experiment with the ”private detection” protocol on the MOT Challenge named the method "MOT\_FCG\_plus" and obtained competitive experimental results. Due to the limited number of submission attempts on the MOT Challenge test, we evaluate on the MOT17 and MOT20 datasets under the "private detection" protocol, using the first half of each video in the training set of MOT17 for training and the second half as the validation set\cite{Philipp2020Tracking}. We provide a brief overview of the some important components used. Namely, for the detection portion of the objects, we fine-tune the YOLOX detector \cite{2021YOLOX} used in MOT\_FCG, and we extract ReID features using the SBS network \cite{10.1145/3581783.3613460} trained on Market1501 \cite{2015Scalable}. The sliding window length is fixed at 6, and the detector threshold is set at 0.7. For different datasets, we fine-tune hyperparameters $\beta_{f}$ and $off$:  $\beta_{f}=0.822, off=0.525$ for MOT17; $\beta_{f}=0.66, off=0.9$ for MOT20; and $\beta_{f}=0.8, off=0.1$ for DanceTrack.

\subsection{Benchmark Results}
In this section, we present the benchmark results of the algorithm on the MOT17 and MOT20 sets. Notably, the modifications we propose generally outperform MOT\_FCG in both datasets. In the table, bold black indicates first place, bold blue indicates second place, and bold red indicates third place.

\subsubsection{Compare with MOT\_FCG}

We present the performance of MOT\_FCG++ on the MOT17-test set and MOT20-test set in Table \ref{tab:w1}. It is worth noting that our algorithm achieves SOTA level in FP, Prcn and DetPr metrics, ranking first in the private detection rankings. Compared to the baseline MOT\_FCG, MOT\_FCG++ shows improvements across all MOT metrics.This shows that compared to the original algorithm, our improvements in the representation based on spatio-temporal motion and appearance features allow for a more precise depiction of similarity relationships among targets. They also provide greater robustness and globality in representing appearance features for long trajectories. Consequently, there are significant and consistent performance improvements across related metrics. We also added StrongSORT's track post-processing method (AFlink+GSI) \cite{
du2023strongsort}, and the results show that the performance of the algorithm is further improved as expected.

\begin{table}[htbp]
\centering
\caption{\label{tab:w1}Compare with MOT\_FCG on MOT17-test and MOT20-test. $\uparrow$ indicates that a higher value represents better performance, while $\downarrow$ indicates that a lower value represents better performance. Bold black indicates the best. The addition of * represents experimental results using trajectory post-processing. }
\resizebox{\linewidth}{!}{
\begin{tabular}{c||ccccc|ccccc}
\Xhline{2pt}
\multirow{2}*{\textbf{Tracker}} & \multicolumn{5}{c}{\textbf{MOT17}} & \multicolumn{5}{|c}{\textbf{MOT20}} \\
~ & \textbf{HOTA$\uparrow$} & \textbf{DetA$\uparrow$} & \textbf{AssA$\uparrow$} & \textbf{MOTA$\uparrow$} & \textbf{IDF1$\uparrow$} & \textbf{HOTA$\uparrow$} & \textbf{DetA$\uparrow$} & \textbf{AssA$\uparrow$} & \textbf{MOTA$\uparrow$} & \textbf{IDF1$\uparrow$} \\
\Xhline{2pt}
\textbf{MOT\_FCG \cite{girbau2022multiple}} & {62.6} & {62.2} & {63.4} & {76.7} & {77.7} & {57.3} & {56.7} & {58.1} & {68.0} & {69.7} \\

\textbf{MOT\_FCG++(ours)} & {63.1(+0.5)} & {62.2} & {64.2(+0.8)} & {76.9(+0.2)} & {78.2(+0.5)} & {58.4(+1.1)} & {56.7} & {60.3(+2.2)} & {68.1(+0.1)} & {72.3(+2.6)} \\

\textbf{MOT\_FCG++*(ours)} & \textbf{64.1}(+1.5) & \textbf{64.3}(+2.1) & \textbf{64.3}(+0.9) & \textbf{79.2}(+2.5) & \textbf{79.0}(+1.3) & \textbf{-} & \textbf{-} & \textbf{-} & \textbf{-} & \textbf{-} \\

\Xhline{2pt}  
\end{tabular}
}  
\end{table}

\subsubsection{Compare with other algorithm}
\textbf{MOT17.}  In comparison with current SOTA algorithms, MOT\_FCG++ achieves a MOTA of 80.4, ranking first in the list. In terms of identity association evaluation, MOT\_FCG++ achieves the optimal IDF1 of 81.3. In the HOTA metric, which integrates detection, association and localization effects, MOT\_FCG++ also performs well with a score of 76.1. This indicates that our algorithm modifications contribute to enhancing the competitiveness of MOT\_FCG, demonstrating that our tracker is robust and effective across different scenarios.

\begin{table}[htbp] 
\centering 
\caption{\label{tab:widget}Results on MOT17-val. $\uparrow$ indicates that a higher value represents better performance, while $\downarrow$ indicates that a lower value represents better performance. Bold black indicates first place, bold blue indicates second place, and bold red indicates third place.}

\resizebox{\linewidth}{!}{
\begin{tabular}{c||ccccccccccc}
\Xhline{2pt} 
\textbf{Tracker} & \textbf{HOTA$\uparrow$} & \textbf{DeTA$\uparrow$} & \textbf{AssA$\uparrow$} & \textbf{MOTA$\uparrow$} & \textbf{IDF1$\uparrow$} & \textbf{IDP$\uparrow$} & \textbf{IDR$\uparrow$} & \textbf{IDSW$\downarrow$} & \textbf{Rcll$\uparrow$} & \textbf{FP$\downarrow$} & \textbf{FN$\downarrow$}\\\hline  

\Xhline{1pt}
{SORT \cite{bewley2016simple}} & {69.6} & {73.5} & {66.5} & {74.5} & {77.0} & {83.5} & {71.5} & 867 & 80.3 & 5.2  & 19.7 \\

{DeepSORT \cite{2017Simple}} & {71.0} & {74.8} & {67.8} & {75.3} & 77.3 & {82.3} & {72.8} & 744 & 82.1 & 6.3 & 17.9 \\

{MOTDT \cite{2018Real}} & 71.6 & 75.0 & 68.8 & {\color{red}{\textbf{75.9}}} & {77.8} & 84.0 & {72.5} & 837 & {81.3} & {\color{red}{\textbf{4.9}}} & {18.7} \\

{OC-SORT \cite{cao2023observation}} & 69.3 & 71.4 & 67.4 & 73.9 & 77.8 & {\textbf{87.2}} & 70.2 & {624} & 77.4 & {3.1} & 22.6 \\
 
{UCMCTrack \cite{yi2024ucmctrack}} & {\textbf{77.2}} & 74.1 & {\color{red}{\textbf{74.7}}} & 75.0 & \textbf{82.0} & {\color{red}{\textbf{85.7}}} & {75.4} & {\color{red}{\textbf{525}}} & {\color{blue}{\textbf{83.7}}} & 5.4  & {\textbf{16.1}} \\

{Deep-OC-SORT \cite{2022Observation}} & {69.5} & {71.4} & {67.8} & {73.9} & 77.7 & {\color{blue}{\textbf{87.1}}} & {70.1} & 621 & 77.4 & {\color{blue}{\textbf{3.1}}} & 22.6 \\

{Hybrid-SORT \cite{yang2024hybrid}} & 71.3 & 74.3 & 68.5 & 73.9 & {76.8} & 80.5 & {73.4} & 786 & {82.8} & {8.4} & {\color{red}{\textbf{17.2}}} \\

{Hybrid-SORT-reid \cite{yang2024hybrid}} & 73.0 & 74.5 & 71.7 & 75.2 & 79.7 & 85.0 & 75.0 & {\textbf{462}} & 81.9 & 6.3 & 18.1 \\

{BoT-SORT \cite{aharon2022bot}
} & {\color{blue}{\textbf{76.3}}} & {\textbf{77.2}} & {\color{blue}{\textbf{75.6}}} & {\color{blue}{\textbf{77.2}}} & {\color{blue}{\textbf{81.4}}} & 85.8 & {\color{blue}{\textbf{77.5}}} & {492} & \textbf{83.9} & 6.3  & 18.1 \\

{Tracktor++ \cite{2019Tracking}} & {62.0} & {62.8} & {61.4} & {61.8} & 67.2 & {86.8} & {54.8} & {636} & 62.6 & {\textbf{2.6}} & 37.4 \\

{TransTrack \cite{2020TransTrack}} & 66.2 & 72.1 & 61.2 & 71.2 & {70.4} & 75.1 & {66.2} & 1122 & {80.0} & {8.1} & {20.0} \\

{ByteTrack \cite{2022Bytetrack}} & 72.6 & {\color{blue}{\textbf{75.8}}} & 69.7 & 76.3 & 79.3 & 83.0 & 75.1 & {\color{blue}{\textbf{509}}} & {\color{red}{\textbf{83.0}}} & 6.4 & {\color{blue}{\textbf{17.0}}} \\
 
\Xhline{2pt} 
\textbf{MOT\_FCG} \cite{girbau2022multiple} & 74.7 & {75.4} & 74.4 & {80.4} & 80.5 & 84.4 & {\color{red}{\textbf{76.5}}} & {534} & {80.8} & 5.5 & 19.2 \\

\textbf{MOT\_FCG++(ours)} & {\color{red}{\textbf{76.1}}} & {\color{red}{\textbf{75.4}}} & {\textbf{77.1}} & {\textbf{80.4}} & {\color{red}{\textbf{81.3}}} & {85.3} & {\textbf{77.7}} & {539} & 80.8 & 5.5 & 19.2 \\ 

\Xhline{2pt}
\end{tabular} 
}
\end{table}

\textbf{MOT20.} Compared to MOT17, MOT20 presents more congestion and occlusion scenarios, posing greater challenges to the performance of trackers. We present the performance of MOT\_FCG++ on the MOT20 validation set in Table \ref{tab:widgets}. Compared to the baseline MOT\_FCG, the HOTA of MOT\_FCG++ improves from 65.7 to 65.8, and IDF1 increases from 72.2 to 72.5, indicating a significant enhancement in tracking performance under congested and occluded conditions. In comparison with other SOTA algorithms, the competitiveness of MOT\_FCG++ is not as pronounced as it was on MOT17. This may be attributed to the fact that, although MOT\_FCG++ improves the representation capability of trajectory clustering appearance features, it is ultimately based on adjacent appearance feature association methods, which are significantly affected in congested and occluded scenarios.
\begin{table}[htbp] 
\centering
\caption{\label{tab:widgets}Results on MOT20-val.$\uparrow$ indicates that a higher value represents better performance, while $\downarrow$ indicates that a lower value represents better performance. Bold black indicates first place, bold blue indicates second place, and bold red indicates third place.}

\resizebox{\linewidth}{!}{
\begin{tabular}{c||ccccccccccc}
\Xhline{2pt}
\textbf{Tracker} & \textbf{HOTA$\uparrow$} & \textbf{DeTA$\uparrow$} & \textbf{AssA$\uparrow$} & \textbf{MOTA$\uparrow$} & \textbf{IDF1$\uparrow$} & \textbf{IDP$\uparrow$} & \textbf{IDR$\uparrow$} & \textbf{IDSW$\downarrow$} & \textbf{Rcll$\uparrow$} & \textbf{FP$\downarrow$} & \textbf{FN$\downarrow$}\\\hline 

\Xhline{1pt} 

{SORT \cite{bewley2016simple}} & {66.9} & {68.0} & {\color{red}{\textbf{65.9}}} & {62.8} & {70.8} & {\color{red}{\textbf{76.4}}} & 65.9 & 1235 & 74.6 & 11.6  & 25.4 \\

{DeepSORT \cite{2017Simple}} & {63.4} & {68.9} & {58.6} & {63.6} & 65.1 & {69.7} & {61.0} & 2026 & 75.7 & 11.8 & 24.3 \\

{MOTDT \cite{2018Real}} & 61.4 & 66.8 & 56.7 & 61.4 & {64.1} & 70.8 & {58.5} & 2221 & {72.2} & 10.4 & {27.8} \\ 

{OC-SORT \cite{cao2023observation}} & 65.6 & 68.0 & 63.5 & 62.7 & 69.8 & 75.3 & 65.0 & {1420} & 74.6 & 11.7 & 25.4 \\

{UCMCTrack \cite{yi2024ucmctrack}} & {\color{blue}{\textbf{68.8}}} & {\color{blue}{\textbf{74.2}}} & {\textbf{73.8}} & 64.3 & 71.3 & 71.7 & {\color{blue}{\textbf{70.8}}} & {\color{blue}{\textbf{1142}}} & {\color{blue}{\textbf{86.7}}} & 10.4  & {\color{red}{\textbf{20.7}}} \\

{Deep-OC-SORT \cite{2022Observation}} & {66.1} & {68.1} & {64.3} & {63.0} & 70.3 & 76.0 & {65.4} & 1392 & 74.7 & 11.4 & 25.3 \\

{Hybrid-SORT \cite{yang2024hybrid}} & 68.9 & {\color{red}{\textbf{71.4}}} & 66.7 & 66.4 & {\color{blue}{\textbf{73.0}}} & 75.1 & {\textbf{71.0}} & 1310 & {\color{red}{\textbf{80.5}}} & {14.0} & {\color{blue}{\textbf{19.5}}} \\ 

{Hybrid-SORT-reid \cite{yang2024hybrid}} & {\textbf{70.2}} & 70.2 & {\color{blue}{\textbf{70.4}}} & 65.5 & {\textbf{74.5}} & {\color{blue}{\textbf{78.4}}} & {\textbf{71.0}} & \textbf{889} & 78.1 & 12.5 & 21.9 \\

{BoT-SORT \cite{aharon2022bot}} & {\color{red}{\textbf{67.9}}} & 70.4 & 65.9 & 65.4 & 70.6 & 74.6 & {67.1} & {\color{red}{\textbf{1274}}} & {77.8} & 12.2  & {22.2} \\

{Tracktor++ \cite{2019Tracking}} & {57.8} & {64.9} & {51.6} & {66.5} & 58.4 & {70.7} & {49.8} & {3021} & 68.6 & {11.9} & 31.4 \\

{TransTrack \cite{2020TransTrack}} & 67.4 & \textbf{80.2} & 57.2 & \textbf{78.7} & {69.7} & 71.4 & {\color{red}{\textbf{68.0}}} & 3162 & \textbf{87.2} & \textbf{8.0} & \textbf{12.8} \\

{ByteTrack \cite{2022Bytetrack}} & {66.5} & {70.1} & {63.4} & {65.2} & 69.2 & 73.2 & 65.6 & {1334} & 77.5 & 12.1 & 22.5 \\

\Xhline{2pt}  
\textbf{MOT\_FCG} \cite{girbau2022multiple} & 65.7 & 67.2 & 64.3 & {\color{red}{\textbf{68.1}}} & 72.2 & {\textbf{78.8}} & {60.8} & {1482} & 68.4 & {\color{red}{\textbf{10.3}}} & 31.6 \\

\textbf{MOT\_FCG++(ours)} & 65.8 & 67.2 & 64.5 & {\color{blue}{\textbf{68.3}}} & {\color{red}{\textbf{72.5}}} & {\textbf{78.8}} & 61.2 & {1572} & {68.4} & {\color{blue}{\textbf{10.1}}} & 25.7 \\

\Xhline{2pt}
\end{tabular} 
} 
\end{table}
 
\FloatBarrier 
\subsection{Ablation Study}
\begin{figure}[htbp]
\centering
\includegraphics[width=0.9\linewidth]{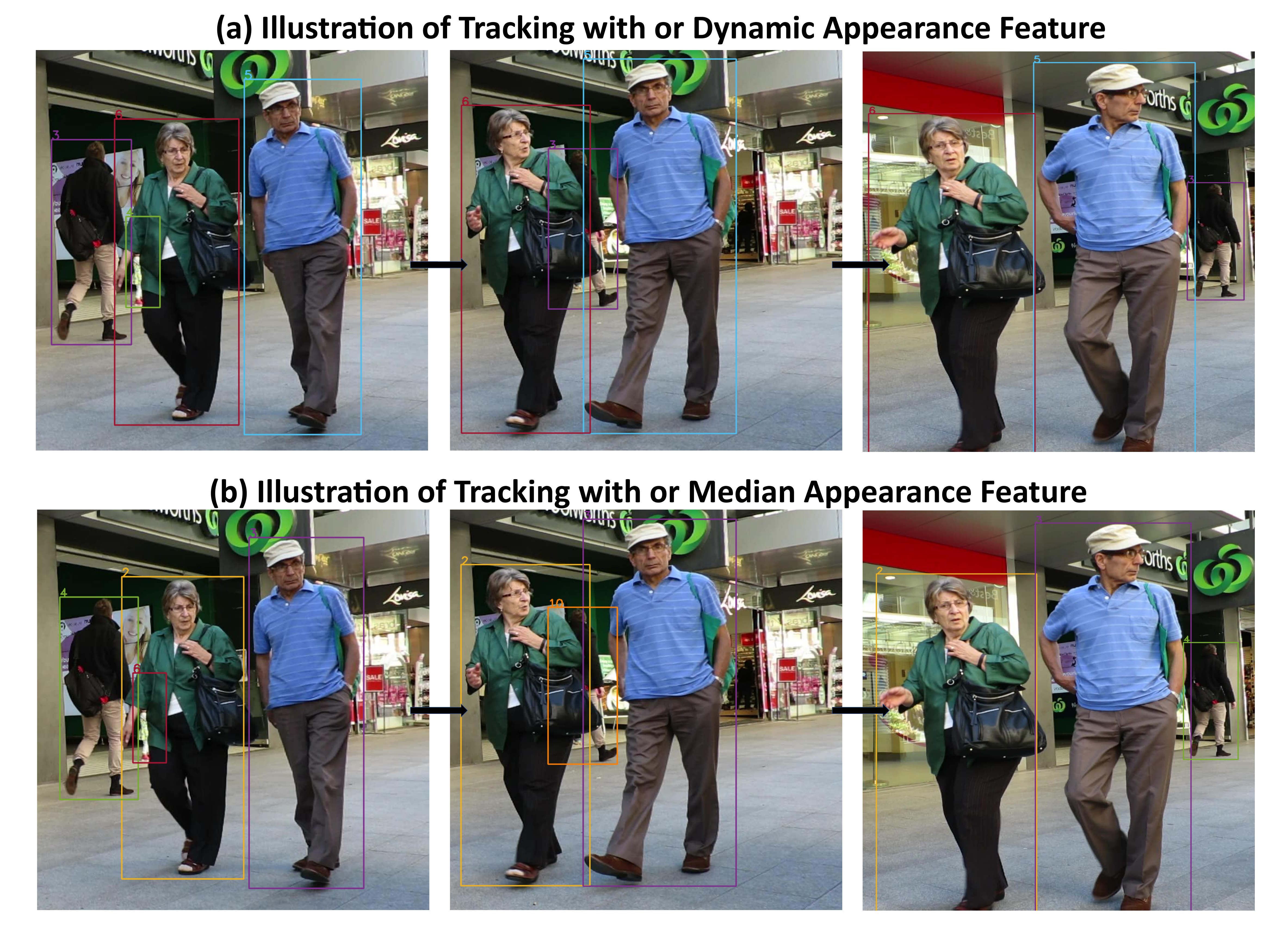} 
\caption{\label{fig:Ablation}Comparison of Different Appearance Feature Representations.} 
\end{figure}

\textbf{Global ablation.} To validate the effectiveness of the proposed modules, we conduct ablation experiments on MOT17-val, MOT20-val, and DanceTrack-val set, using MOT\_FCG as the baseline and sequentially adding Dynamic Appearance(DA), Diagonal Modulated GIoU(DGIoU) and  Average Constant Velocity Modeling(ACV). The results of the ablation experiments are shown in Table \ref{tab:xiao}. We visualize the tracking results, as shown in Figure \ref{fig:Ablation}. Figure (a) uses the median element of the original MOT\_FCG as the appearance feature for the trajectory, resulting in No.4 switching identity to N0.10 during occlusion. In contrast, Figure (b) employs the dynamic appearance feature representation proposed in this paper, which maintains identity consistently during occlusion. Therefore, DA is capable of providing a global representation of the appearance embedding features for trajectory clustering, which prevents identity switching during occlusions. From Table \ref{tab:xiao}, it can be seen that DA performs better on MOT17 and DanceTrack. However, in the dense and highly occluded scenarios of MOT20, HOTA show limited improvements, possibly due to frequent target occlusions restricting the extraction of global features. DGIoU shows overall improvements across all three datasets, indicating its superior representation of target spatial location information compared to IoU. Regarding ACV, although its impact on tracking  performance is relatively small, it does suggest that it is more robust and competitive compared to the constant velocity model.

\begin{table}[htbp]
\centering
\caption{\label{tab:xiao}Ablation study on MOT17-val, MOT20-val and DanceTrack-val set. $\uparrow$ indicates that a higher value represents better performance, while $\downarrow$ indicates that a lower value represents better performance. Bold black indicates the best.}
\resizebox{\linewidth}{!}{
\begin{tabular}{ccc||cccc|cccc|cccc}
\Xhline{2pt}
\multirow{2}*{\textbf{DA}} & \multirow{2}*{\textbf{DGIoU}} & \multirow{2}*{\textbf{ACV}} & \multicolumn{4}{c}{\textbf{MOT17}} & \multicolumn{4}{|c|}{\textbf{MOT20}} & \multicolumn{4}{c}{\textbf{DanceTrack}} \\
~ & ~ & ~ & \textbf{HOTA$\uparrow$} & \textbf{MOTA$\uparrow$} & \textbf{IDF1$\uparrow$} & \textbf{IDSW$\downarrow$} & \textbf{HOTA$\uparrow$} & \textbf{MOTA$\uparrow$} & \textbf{IDF1$\uparrow$} & \textbf{IDSW$\downarrow$} & \textbf{HOTA$\uparrow$} & \textbf{MOTA$\uparrow$} & \textbf{IDF1$\uparrow$} & \textbf{IDSW$\downarrow$}  \\
\Xhline{2pt}
~ & ~ & ~ & {74.73} & {80.36} & {80.54} & {534} & {65.71} & 68.11 & {72.18} & 1482 & {52.93} & {53.52} & {74.01} & {1835} \\
{\Checkmark} & ~ & ~ & {75.22} & {80.38} & {80.67} & \textbf{527} & {65.62} & {68.25} & {72.21} & \textbf{1469} & {53.12} & {53.88} & \textbf{74.15} & {1631}\\
{\Checkmark} & {\Checkmark} & ~ & {76.09} & {80.43} & {81.07} & {541} & {65.76} & \textbf{68.31} & {72.40} & {1558} & {53.83} & {54.31} & {74.13} & {1696}\\
{\Checkmark} & {\Checkmark} & {\Checkmark} & \textbf{76.14} & \textbf{80.43} & \textbf{81.32} & {539} & \textbf{68.81} & {68.30} & \textbf{72.53} & {1572} & \textbf{54.15} & \textbf{54.62} & 74.12 & \textbf{1582} \\ 
\Xhline{2pt} 
\end{tabular} 
}
\end{table}
  
\textbf{Selection of DGIoU.} We perform ablation experiments on similarity metrics that reflect spatial location relationships. Similar to DGIoU, WGIoU is modulated by using width on top of GIoU, and HGIoU uses height to replace width. The results are shown in the table \ref{tab:widgets1} below, compared to the first row, GIoU improves the HOTA by 0.497 due to the fact that GIoU better reflects the spatial location relationship between targets. By introducing different modulation terms to GIoU, we found that using diagonal modulation is the best, the reason is that compared with width and height modulation, diagonal modulation combines the width and height information, which better reflects the scale size and relative position of the targets.
\begin{table}[htbp] 
\centering
\caption{\label{tab:widgets1}Results of different spatial similarity metrics on the MOT17-train.}

\resizebox{0.7\linewidth}{!}{
\begin{tabular}{c||cccccc}
\Xhline{2pt}
~ & \textbf{HOTA$\uparrow$} & \textbf{MOTA$\uparrow$} & \textbf{IDF1$\uparrow$} & \textbf{DeTA$\uparrow$} & \textbf{AssA$\uparrow$} & \textbf{IDSW$\downarrow$} \\\hline

\Xhline{1pt} 
\textbf{IoU} & {77.06} & {87.19} & {86.09} & {77.67} & 76.76 & 2262 \\
\textbf{GIoU} & {77.55} & \textbf{87.34} & {86.94} & {77.65} & 77.76 & \textbf{1767} \\
\textbf{WGIoU} & {77.05} & {87.26} & {85.96} & {77.66} & 76.76 & 2025 \\
\textbf{HGIoU} & {77.49} & {87.30} & {86.58} & {77.69} & 77.59 & 1890 \\
\textbf{DGIoU} & \textbf{77.70} & {87.30} & \textbf{87.06} & \textbf{77.74} & \textbf{77.96} & 1905 \\

\Xhline{2pt}
\end{tabular} 
} 
\end{table}

\begin{figure}[htbp]
\centering
\includegraphics[width=0.9\linewidth]{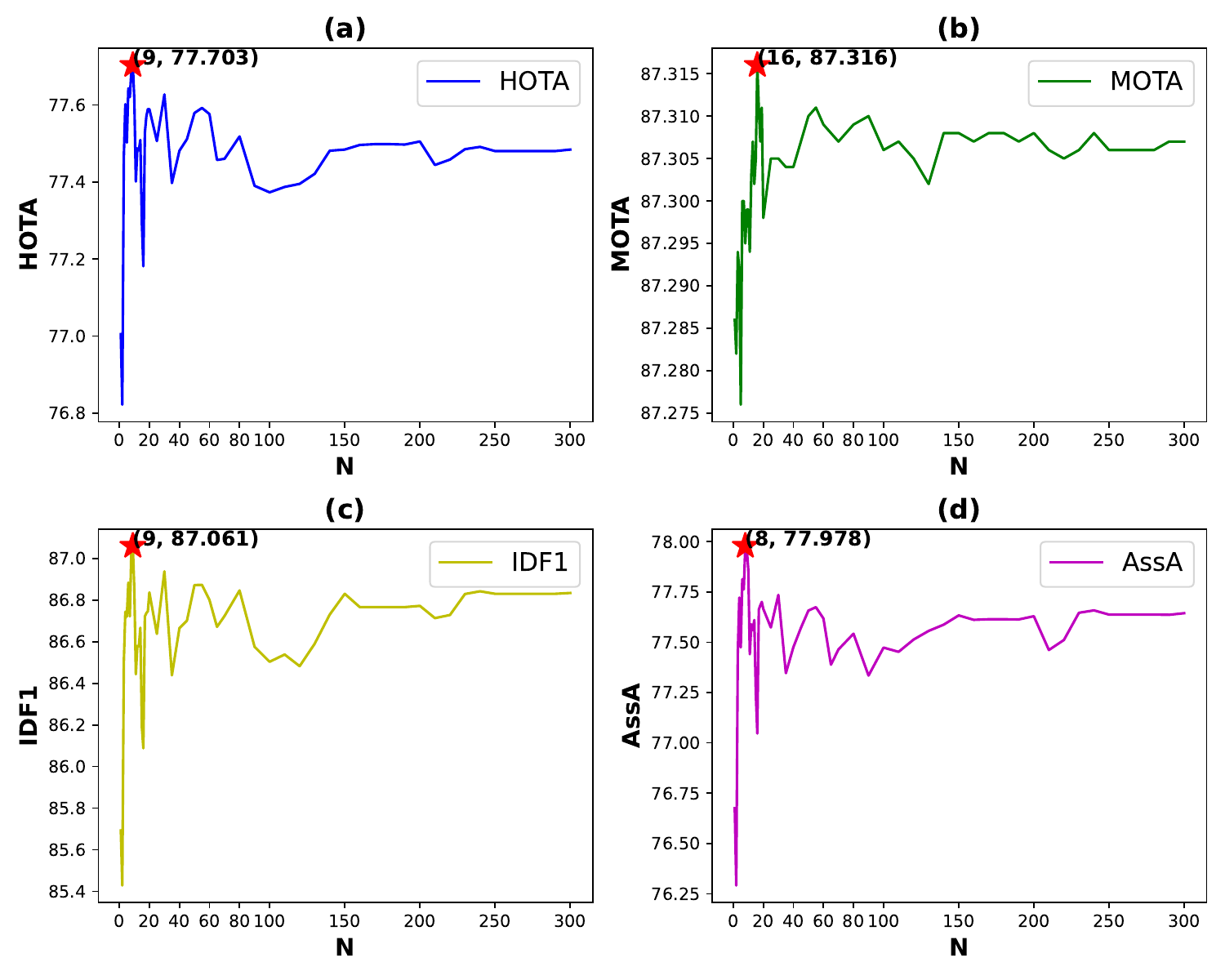} 
\caption{\label{fig:N_Ablation}Tracking performance of MOT\_FCG++ on the MOT17-train with different N values.} 
\end{figure}

\textbf{Selection of N.} In the figure \ref{fig:N_Ablation}, we show the effect of different values of $N$ for ACV on the tracker performance on the MOT17 set. In the experiments, $N=2$ represents the motion modeling approach of MOT\_FCG using the last two frame spacing as the velocity estimation, and $N>2$ represents the ACV we use. For the MOT17 benchmark experiments, we can see that the performance of using the ACV is significantly better than the original one when $N>2$, which is due to the use of the average velocity modeling to effectively filter out the observation noise, and is more stable than the original one. The oscillation frequency of the curve is larger when $N$ is small, and the tracking performance gradually converges and stabilizes as $N$ increases. From the results, it can be seen that the tracking performance is better when $N=9$, so we take $N=9$ for the experiments.

\textbf{Selection of DA.} We conducted ablation experiments on the representation methods for trajectory appearance features. In this context, "Median App." refers to the median element feature representation used by MOT\_FCG, "Max App." denotes the representation using the appearance feature with the highest confidence, and "Mean App." represents the average appearance feature representation of the trajectory observations. "Dynamic App." is the feature representation method proposed in this paper.From the results presented in the table \ref{tab:widgets2}, it is evident that using Dynamic App. yields the best tracking performance, with a HOTA improvement of 0.234 compared to the first row. The Max App. approach focuses solely on the impact of detection quality on the representation method, and the Mean App. approach concentrates exclusively on global information of the trajectory, both of which have inherent limitations. Therefore, Dynamic App. effectively balances detection quality and global information, making it a more robust and accurate approach to appearance feature representation.
\begin{table}[htbp] 
\centering
\caption{\label{tab:widgets2}Results of different appearance feature representation on the MOT17-train.}

\resizebox{0.7\linewidth}{!}{
\begin{tabular}{c||cccccc}
\Xhline{2pt}
~ & \textbf{HOTA$\uparrow$} & \textbf{MOTA$\uparrow$} & \textbf{IDF1$\uparrow$} & \textbf{DeTA$\uparrow$} & \textbf{AssA$\uparrow$} & \textbf{IDSW$\downarrow$} \\\hline

\Xhline{1pt} 
\textbf{Median App.} & {77.47} & \textbf{87.35} & {86.74} & {77.66} & 77.58 & \textbf{1710} \\
\textbf{Max App.} & {70.66} & {86.96} & {75.49} & {77.87} & 64.48 & {1875} \\
\textbf{Mean App.} & {75.76} & {87.29} & {84.54} & {77.62} & 74.25 & 1875 \\
\textbf{Dynamic App.} & \textbf{77.70} & {87.30} & \textbf{87.06} & \textbf{77.74} & \textbf{77.96} & 1905 \\

\Xhline{2pt}
\end{tabular} 
} 
\end{table}

\FloatBarrier
\section{Conclusion}
In this paper, we propose an enhanced and effective MOT method - MOT\_FCG++, based on the hierarchical clustering MOT method MOT\_FCG. Compared to MOT\_FCG, MOT\_FCG++ utilizes a dynamic appearance embedding representation that incorporates more global appearance and confidence information. Besides, the proposed Diagonal Modulated GIoU also demonstrates greater robustness and generalization in spatial motion feature representation. The introduction of Average Constant Velocity Modeling also leads to a slight improvement in tracking performance. MOT\_FCG++ achieves excellent performance and competitive results on the MOT17-sets, MOT20-sets, and DanceTrack-sets. We hope that these improved methods will inspire future representations of motion and appearance cues in clustering MOT methods.
 
\newpage

\bibliographystyle{unsrt}
\bibliography{sample}

\end{document}